\documentclass[letterpaper]{article} 
\usepackage{aaai24}  
\usepackage{times}  
\usepackage{helvet}  
\usepackage{courier}  
\usepackage[hyphens]{url}  
\usepackage{graphicx} 
\usepackage{natbib}  
\usepackage{caption} 
\usepackage{graphicx}
\usepackage{subfigure}
\usepackage{amsmath}
\usepackage{amssymb}
\usepackage{booktabs}
\usepackage{float}
\usepackage[pagebackref,breaklinks,colorlinks,bookmarks=false]{hyperref}

\usepackage[capitalize]{cleveref}
\usepackage{hyperref}
\usepackage{xcolor}

\usepackage{caption}

\usepackage{algorithm}
\usepackage{algorithmic}
\usepackage{graphicx}
\usepackage{amsmath}
\usepackage{amssymb}
\usepackage{ragged2e} 
\usepackage{booktabs,makecell, multirow, tabularx}
\usepackage{multirow}
\usepackage[capitalize]{cleveref}
\crefname{section}{Sec.}{Secs.}
\Crefname{section}{Section}{Sections}
\crefname{table}{Table}{Tabs.}
\crefname{figure}{Figure}{Fig.}
\usepackage{newfloat}
\usepackage{listings}
\DeclareCaptionStyle{ruled}{labelfont=normalfont,labelsep=colon,strut=off} 
\floatstyle{ruled}
\newfloat{listing}{tb}{lst}{}
\floatname{listing}{Listing}
\title{UVOSAM: A Mask-free Paradigm for Unsupervised Video Object Segmentation via Segment Anything Model}
\author{
Zhenghao Zhang, Shengfan Zhang, Zuozhuo Dai, Zilong Dong, Siyu Zhu
}
\affiliations{Alibaba Group}
\usepackage{bibentry}

\begin{document}

\maketitle

\begin{abstract}

The current state-of-the-art methods for unsupervised video object segmentation (UVOS) require extensive training on video datasets with mask annotations, limiting their effectiveness in handling challenging scenarios. However, the Segment Anything Model (SAM) introduces a new prompt-driven paradigm for image segmentation, offering new possibilities. In this study, we investigate SAM's potential for UVOS through different prompt strategies. We then propose UVOSAM, a mask-free paradigm for UVOS that utilizes the STD-Net tracker. STD-Net incorporates a spatial-temporal decoupled deformable attention mechanism to establish an effective correlation between intra- and inter-frame features, remarkably enhancing the quality of box prompts in complex video scenes. Extensive experiments on the DAVIS2017-unsupervised and YoutubeVIS19\&21 datasets demonstrate the superior performance of UVOSAM without mask supervision compared to existing mask-supervised methods, as well as its ability to generalize to weakly-annotated video datasets. Code can be found at \href{https://github.com/alibaba/UVOSAM}{https://github.com/alibaba/UVOSAM}.

\end{abstract}

\section{Introduction}

Video object segmentation (VOS)~\cite{miao2020memory,zhou2022survey,yang2020collaborative} is a crucial computer vision task that aims to segment primary objects in video sequences. 
It has wide-ranging applications in video editing, autonomous driving, and robotics. 
VOS can be classified into two main types: semi-supervised VOS~\cite{oh2019video,xie2021efficient, chen2019multilevel, park2022per}, which involves supervision in both training and inference, and unsupervised VOS~\cite{liu2021f2net,cho2023treating, lee2023unsupervised, ren2021reciprocal, yang2021learning}, which generally require no prior data during inference.
\begin{figure}[t] 
\centering 
\includegraphics[scale=0.45]{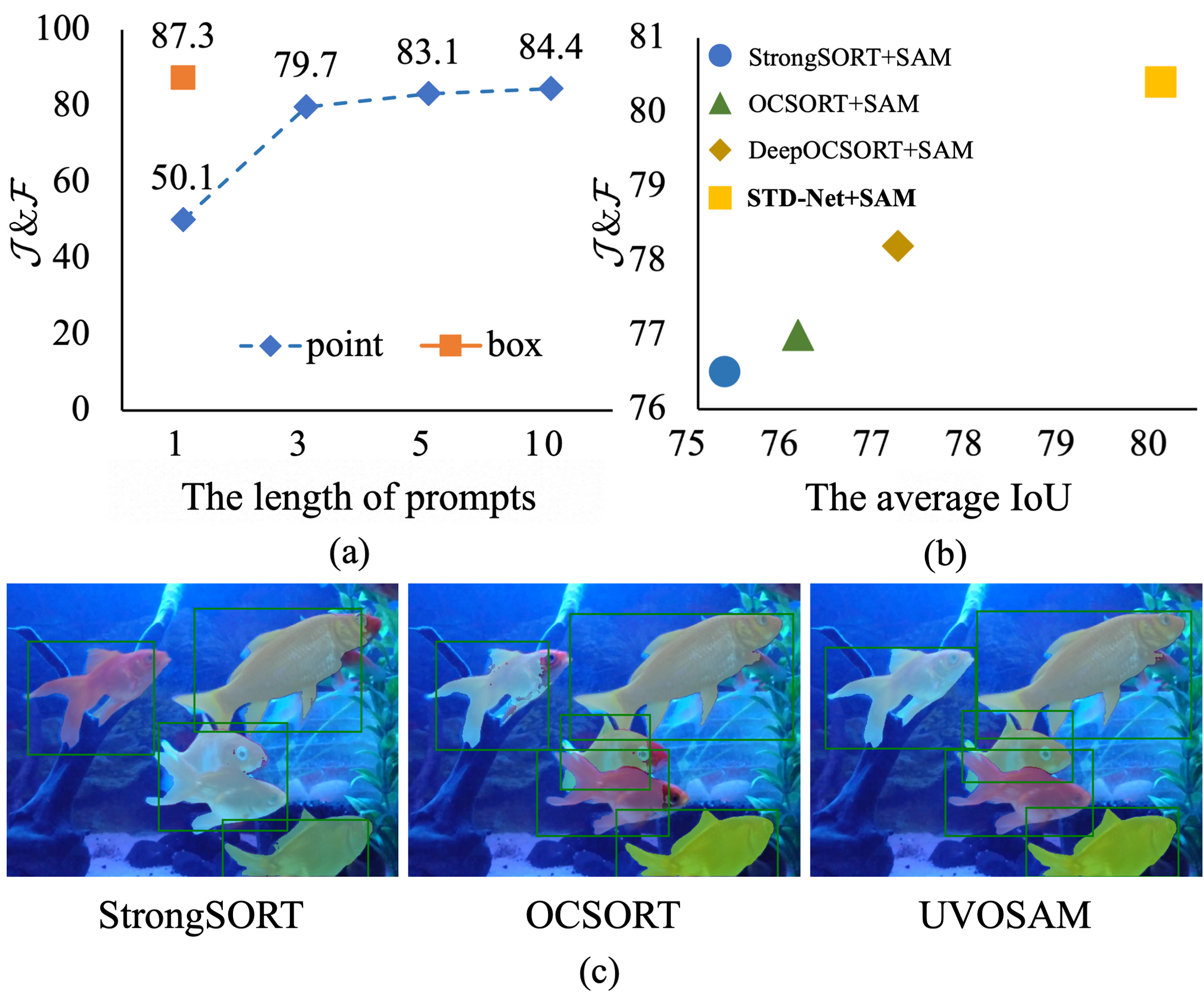} 
\caption{
(a) The plot depicts the $\mathcal{J}\&\mathcal{F}$ scores against the length of prompts on the DAVIS2017-unsupervised~\cite{pont20172017}. 
For box prompts, we provide ground-truth boxes with IDs. 
As for point prompts, we randomly sample a varying number of points from the ground-truth masks. 
The median of 5 runs is reported, along with the standard deviation.
(b) The performance of box prompts generated by tracking models is evaluated. 
The average IoU is computed following the methodology ~\cite{rezatofighi2019generalized}.
(c) An illustration showcasing the ambiguous issue caused by unsuitable shapes and locations of box prompts. It is recommended to view the image in a zoomed-in way for clarity.
} 
\label{fig1} 
\vspace{-3mm}
\end{figure}

Semi-supervised VOS leverages the initial segmentation mask from the first frame to initialize the model, enabling it to track and segment specific objects throughout the entire video sequence. On the other hand, unsupervised VOS (UVOS) relies on the model to autonomously discover and extract masks for prominent objects without any prior information. UVOS is becoming popular for its automatic segmentation capability during inference, eliminating the need for manual annotation. However, current state-of-the-art UVOS methods~\cite{cho2023treating, lee2023unsupervised, pei2022hierarchical} still require a large-scale manually annotated video dataset~\cite{perazzi2016benchmark, xu2018large, qi2022occluded, ding2023mose} for training. Annotating video object segmentation is prohibitively expensive as it involves providing masks and trajectories. Even annotating masks using coarse polygons is significantly more time-consuming than annotating video bounding boxes~\cite{cheng2022pointly}. The high cost of annotating masks poses a challenge in scaling existing VOS approaches. Therefore, a mask-free setting may be necessary for more efficient handling of this task.


Recent advancements in segmentation foundation models~\cite{kirillov2023segment, wang2023seggpt, zou2023segment} have greatly enhanced image segmentation.
One notable model in this field is the Segment Anything Model (SAM)~\cite{kirillov2023segment}, which has emerged as a leading contender.
SAM's training on a vast dataset of over 1 billion masks has enabled it to achieve impressive zero-shot generalization capabilities. Moreover, SAM is designed to be promptable with various types of prompts, such as bounding boxes and points, enabling it to transfer zero-shot knowledge to new image distributions and tasks. It consistently exhibits robust performance across various downstream tasks, surpassing existing methods in zero-shot settings. This raises an intriguing question: Can we leverage SAM in UVOS by utilizing auto-generated prompts, thereby eliminating the labor-intensive process of segmentation mask annotation?



To achieve our objective, we begin by investigating the most suitable type of prompts for the UVOS task. As shown in~\cref{fig1}(a), even when the number of sampling points is increased to 10, the performance of the bounding box remains superior to that of points. This can be explained intuitively in the following ways. Firstly, the bi-directional cross-attention module in the decoding path heavily relies on the positional encoding of the point prompts, which depends on their coordinates. As the image embeddings are updated accordingly, point prompts located at different positions, despite having similar semantic contexts, may result in variations in the final segmentation masks. Secondly, since SAM is trained for general segmentation rather than specific tasks, it may struggle to accurately handle segmentation boundaries, especially when only a few points are used.
When selecting the bounding box as the prompt, both detection and association are crucial. 
Recent advancements in multi-object tracking (MOT)~\cite{du2023strongsort,cao2023observation,maggiolino2023deep} have shown superior performance in identity association. 
Therefore, our focus is primarily on the detection quality in terms of Intersection over Union (IoU) between the predicted location and the object's boundary. 
As shown in~\cref{fig1}(c), unsuitable shapes and locations of box prompts can lead to segmentation ambiguity issues for SAM.
To address this, we propose STD-Net, a tracking network that combines DeformableDETR~\cite{zhu2020deformable} and DeepOCSORT~\cite{maggiolino2023deep}. In the deformable transformer, we introduce a novel approach called spatial-temporal decoupled deformable attention (STD-DA). This approach allows us to independently leverage multi-scale deformable attention in the spatial and temporal domains, resulting in spatially discriminative and temporally representative features. We also employ a dynamic fusion module to effectively combine these features and enhance the accuracy of non-overlapping bounding boxes in complex video scenes.
As demonstrated in~\cref{fig1}(b), the bounding box prompts generated by STD-Net exhibit significantly superior performance in terms of average IoU compared to state-of-the-art MOT methods. Furthermore, it can be proven that the value of IoU greatly affects the segmentation performance.

In summary, we present UVOSAM, a mask-free paradigm for UVOS.
UVOSAM combines STD-Net and SAM, with STD-Net trained exclusively using bounding box annotations, while SAM is kept frozen to preserve its valuable pre-trained knowledge. 
Our main contributions can be summarized as follows:
\begin{itemize}
    \item We conduct a pioneering study to investigate the potential of SAM for UVOS through the exploration of various prompt strategies. Compared to existing UVOS methods, SAM demonstrates effective segmentation capabilities for weakly-annotated video data.
    \item We propose UVOSAM, an innovative mask-free paradigm for UVOS. UVOSAM employs a novel tracker called STD-Net to generate high-quality trajectories, which are served as box prompts for SAM. STD-Net incorporates a spatial-temporal decoupled deformable attention mechanism to establish a strong correlation between intra- and inter-frame features. This enhancement obviously improves the quality of box prompts, particularly in complex video scenes.
    \item Extensive experiments conducted on the DAVIS2017-unsupervised and YoutubeVIS19\&21 datasets showcase the promising results of UVOSAM and its superiority over existing mask-supervised methods.
\end{itemize}


\section{Related Works}
\subsection{Unsupervised Video Object Segmentation}
UVOS is designed to segment salient objects in a class-agnostic manner, eliminating the need for manual guidance during inference.
This approach reduces the time and cost of Video Object Segmentation (VOS), thus paving the way for a variety of real-time applications.
Early techniques~\cite{fragkiadaki2015learning, tokmakov2017learning, yang2021self} for detecting salient objects assumed that object pixels exhibited the same motion patterns across successive frames, employing motion cues such as optical flow to facilitate the segmentation process. 
However, these methods have a notable limitation: their performance is heavily influenced by the optical flow's quality, especially in situations involving occlusions and rapid motion. 
Several studies~\cite{zhou2020motion, zhang2021deep, ren2021reciprocal, yang2021learning, cho2023treating} have also investigated the use of learning to focus on object appearance features. 
For example, MATNet~\cite{zhou2020motion} employs the MAT-block to convert appearance features into motion-attentive representations, while RTNet~\cite{ren2021reciprocal} mutually transforms appearance and motion features to identify primary objects. 
Moreover, AMC-Net~\cite{yang2021learning} implements a multi-modality co-attention gate to foster a deep collaboration between appearance and motion data for precise segmentation.
DPA~\cite{cho2022} leverages inter-modality attention to densely incorporate context information from both motion and appearance. 
Lastly, PMN~\cite{lee2023unsupervised} efficiently extracts appearance and motion data by deriving superpixel-based component prototypes.

In the realm of unsupervised multiple object segmentation, the majority of methods employ the track-by-detect paradigm.
This involves generating object proposals using instance segmentation models, which are subsequently tracked by re-identification models.
The Target-Aware model~\cite{zhou2021target}, for instance, uses a target-aware adaptive tracking framework to associate proposals across frames, yielding robust matching results.
Similarly, INO~\cite{pan2022n} enhances the association of proposals by incorporating inter-frame consistency through the affinity matrix.
On the other hand, MCMPG~\cite{yuan2023} builds a graph using object proposals generated from several frames around the key frame, and then propagates them to the key frame to reason about key-frame objects.
Unlike these methods that rely on costly annotation-intensive segmentation datasets for training, our mask-free UVOSAM model achieves accurate segmentation by providing simple prompts to the vision foundation model SAM~\cite{kirillov2023segment}.
This approach makes UVOS more accessible.

\subsection{Vision Foundation Models}
Vision foundation models, which have shown remarkable adaptability across a range of tasks, have recently been attracting significant attention.
CLIP~\cite{radford2021learning}, a groundbreaking model trained on web-scale image-text pairs, demonstrates impressive zero-shot capabilities and has been widely implemented in numerous multi-modal tasks for feature alignment. 
Similarly, BLIP~\cite{li2022blip, li2023blip} utilizes a dataset bootstrapped from large-scale noisy data for multi-modal pre-training, enhancing zero-shot performance in text-to-video retrieval tasks.
In a bid to develop a universal detection framework, UniDetector~\cite{wang2023detecting} uses images from a variety of sources and diverse label spaces for training, achieving zero-shot performance that significantly outperforms supervised methods. 

In the realm of segmentation tasks, SAM~\cite{kirillov2023segment}, the first promptable foundation model, has been recognized for its ability to segment any object in any context using various prompts, such as boxes and points.
Models such as SegGPT~\cite{wang2023seggpt} and SEEM~\cite{zou2023segment}, which share a similar idea with SAM, were proposed as generalist models to perform arbitrary segmentation tasks with different prompt types.
Some contemporaneous works have extended SAM to video segmentation tasks.
For instance, TAM~\cite{yang2023track} and SAM-Track~\cite{Cheng2023SegmentAT} utilize cutting-edge mask trackers to interactively segment video objects, while SAM-PT~\cite{Rajivc2023SegmentAM} employs sparse points input for mask generation.
In this paper, our focus is on the UVOS task, and we optimize the box prompts of SAM to significantly reduce the need for intensive video mask annotation. 
Moreover, our approach inherently possesses SAM's robust zero-shot capability, making it suitable for handling weakly labeled video segmentation datasets.

\subsection{Multi-object Tracking}
Multi-object tracking aims to accurately detect multiple objects while providing continuous trajectories for each object. The dominant paradigm in this field is Tracking-By-Detection, which consists of two tasks: detection and association. To better predict the next position of objects, SORT~\cite{bewley2016sort} first uses the Kalman filter~\cite{welch1995Kalmanfilter} and then matches predicted boxes with detected boxes using the Hungarian algorithm~\cite{kuhn1955hungarian}. DeepSORT~\cite{wojke2017deepsort} effectively mitigates occlusion problems by extracting ReID features of bounding boxes with a separate CNN model. TransTrack~\cite{sun2020transtrack} obtains bounding boxes and ReID features through the vision transformer. ByteTrack~\cite{zhang2022bytetrack} adopts YOLOX~\cite{ge2021yolox} as the detector and tracks bounding boxes with scores below the threshold. StrongSORT~\cite{du2023strongsort} designs an appearance-free link model and comprehensively optimizes DeepSORT, achieving a balance between inference speed and accuracy. OCSORT~\cite{cao2023observation} designs a cumulative error correction module and utilizes a learnable motion model to improve tracking performance under occlusion and non-linear motion conditions. DEEPOCSORT~\cite{maggiolino2023deep} adaptively integrates ReID features with OCSORT and achieves promising tracking results. 
Given that the aforementioned approaches have shown outstanding performance in identity association, our proposed UVOSAM leverages the association methods of DeepOCSORT, primarily focusing on enhancing the detection quality of box prompts in terms of Intersection over Union (IoU) between predicted boxes and object boundaries.

\section{Methodology}
\label{3}


This section provides a detailed introduction to our proposed method, UVOSAM, which stands for Unsupervised Video Object Segmentation using SAM.
Our approach does not depend on any video segmentation mask labels.
We begin by revisiting the SAM architecture and then proceed to present the UVOSAM pipeline, highlighting the key features of STD-Net.


\begin{figure*}[ht] 
\centering 
\includegraphics[scale=0.165]{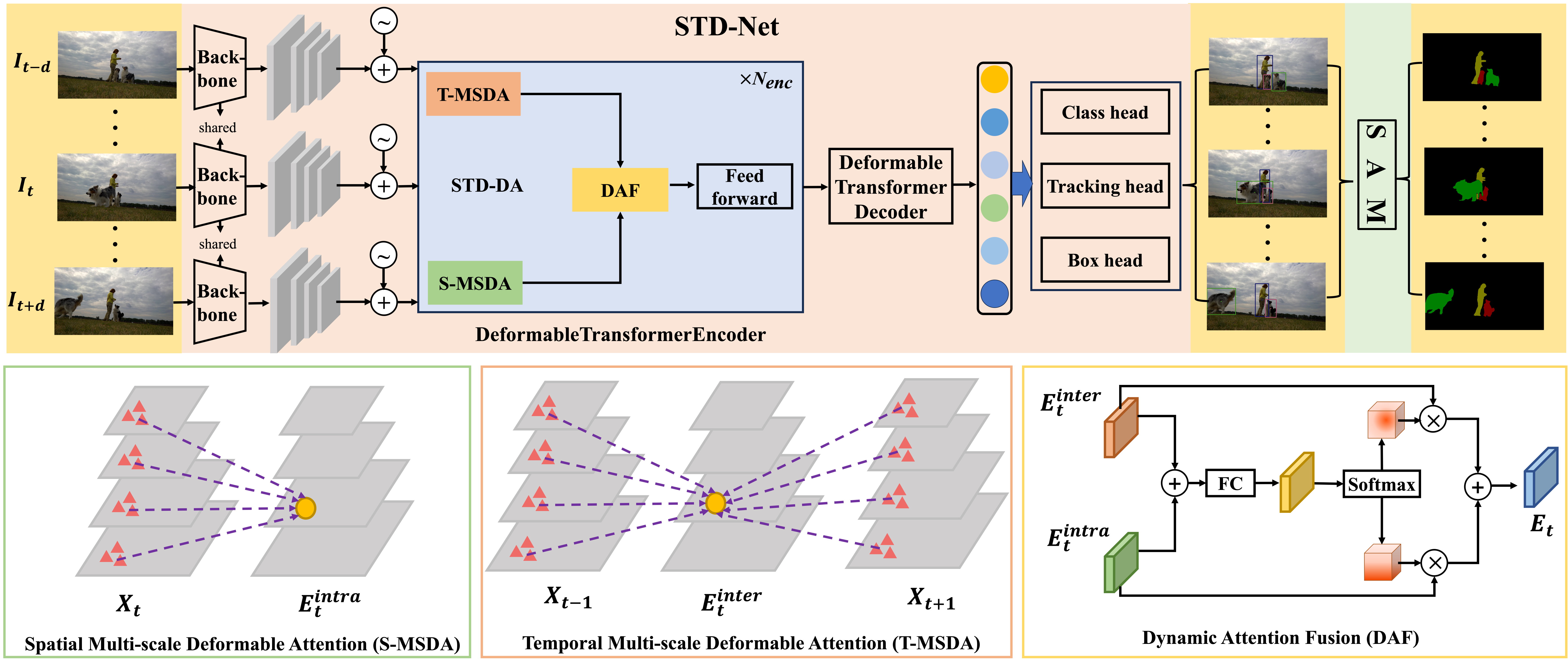} 
\caption{The overall illustration of our proposed UVOSAM, which mainly consists of the STD-Net and SAM.  Given the multi-scale feature maps of a video clip, the transformer encoder first extracts enriched spatial-temporal features via the proposed STD-DA. These features are decoded by object queries, which then generate objectness scores, object locations, and ReID features through three output heads. We adopt the association method proposed in DeepOCSORT to generate identities using ReID features. The boxes with identities serve as prompts for SAM.} 
\label{fig2} 
\end{figure*}

\subsection{Preliminaries: architecture of SAM}
\label{3.1}
SAM is a segmentation foundation model that has been trained using over one billion masks, demonstrating remarkable zero-shot proficiency. Its primary components consist of an image encoder denoted as $E_{i}$, a prompt encoder denoted as $E_{p}$, and a lightweight mask decoder denoted as $D_{m}$. To elaborate further, SAM takes an image $I \in  \mathbb{R}^{H\times W \times 3}$ and its corresponding prompts $P$ (such as points, bounding boxes, text, or masks) as inputs. These two input streams are then embedded separately by the image encoder and prompt encoder, which can be formulated as follows:
\begin{normalsize}
\begin{equation}
\begin{aligned}
F_{i} = E_{i}(I)  \in  \mathbb{R}^{h\times w \times c}  ~~~ F_{p} = E_{p}(P) \in  \mathbb{R}^{k\times c},
\end{aligned}
\end{equation}
\end{normalsize}
where $h$, $w$ and $c$ represent the height, width, and channel number of the feature map, and $k$ denotes the length of the prompt tokens. Following that, $F_{i}$ and $F_{p}$ are passed to $D_{m}$. In addition, there are multiple learnable tokens $F_{t}$ concatenated to $F_{p}$ as a prefix to enable flexible prompting. These two embeddings are then interacted through cross-attention, and the zero-shot mask is generated by decoding the interactive features. This process can be described as follows:
\begin{normalsize}
\begin{equation}
\begin{aligned}
M = D_m(~Attn(F_{i}, concat(F_{p}, F_{t}))~)  \in  \mathbb{R}^{H\times W } 
\end{aligned}
\end{equation}
\end{normalsize}

As mentioned previously, SAM is trained for general segmentation tasks rather than specific ones. Consequently, it may encounter difficulties in accurately handling segmentation boundaries, especially when prompt information is limited. For instance, a single prompt such as a single point prompt can lead to segmentation ambiguity issues, where SAM cannot differentiate which mask the prompt corresponds to. Hence, the objective of this paper is to investigate methods for optimizing the quality of box prompts. By generating high-quality box prompts, we can potentially achieve promising outcomes in UVOS tasks.

\subsection{The overall of proposed UVOSAM}
\label{section3.2}
UVOSAM consists of two components: a STD-Net and a SAM. As shown in  \cref{fig2}, let $\left \{  I_i \right \}_{i=t-d}^{t+d}$ represent the input clip centered at time $t$, which includes the $t$-th frame and its $2d$ neighboring frames. The encoder of STD-Net initially extracts enriched spatial-temporal features $E$ using a novel spatial-temporal decoupled deformable attention (STD-DA) module. These features $E$ are then utilized to refine the object queries and obtain discriminative instance embeddings, which are subsequently decoded into the category, location, and ReID features of each object through three output heads. These outputs are further combined to generate the object trajectories. Subsequently, the object trajectories are utilized as box prompts in SAM to obtain refined video segmentation results. The entire SAM is kept frozen to retain its pre-trained knowledge, and only the STD-Net is trained using category and bounding box annotations.

\subsection{STD-Net} \label{3.3}
Our STD-Net is composed of a vision backbone, a deformable transformer, and three output heads. In order to leverage the temporal information among frames for precise object discovery, we introduce a novel STD-DA module in the transformer encoder to effectively capture spatial-temporal correlations. Furthermore, we employ contrastive learning, following the approach~\cite{wu2022defense}, to obtain ReID features. The key design principles will be explained in detail in the following paragraphs.

\textbf{STD-DA.}  The proposed STD-DA module initially utilizes multi-scale deformable attention separately in the spatial and temporal domains, and subsequently combines them to generate dynamically enriched spatial-temporal features. 
The STD-DA architecture consists of three sub-modules: spatial multi-scale deformable attention (S-MSDA), temporal multi-scale deformable attention (T-MSDA), and dynamic attention fusion.

The S-MSDA module, proposed in this work, calculates attention on points sampled from the multi-scale feature maps of the current frame. Let the $l\text{-}th$ feature map from frame $t$ be denoted as  $x^{l}_t \in  \mathbb{R}^{C\times H_{l} \times W_{l}}$, and the normalized coordinates of the reference point for $q$ as $\hat{p} _{q} \in {[0,1]}^2 $.
With these definitions, the S-MSDA module can be described as follows, given the query feature  $z_{q}$:
\begin{normalsize}
\begin{equation}
\begin{aligned}
\mathcal{S\text{-}MSDA} & (z _{q}, \hat{p} _{q}, x_{t}^{l}) = \sum_{m=1}^{M}W_{m} \cdot [\sum_{l=1}^{L}\sum_{k=1}^{K_{intra}} A_{mlqk}  \\ 
& \cdot W_{m}^{'} x^{l}_t(\phi_{l}(\hat{p} _{q})+\Delta p_{mlqk} ) ], 
\end{aligned}
\end{equation}
\end{normalsize}
where $m$ denotes the index of multiple attention heads, $K_{intra}$ is the total sampled number from each feature level. $\Delta p_{mlqk}$ and $A_{mlqk}$ represent the pixel offset and attention weight of the $k\text{-}th$ sampling point in the $l\text{-}th$ feature level. Both of them are obtained via linear projection over $z_q$. Function $\phi_{l}$ re-scales the normalized coordinates $\hat{p} _{q}$ to the input feature map of the $l\text{-}th$ level.

We further expand the S-MSDA module to the temporal domain and introduce T-MSDA. The key concept behind T-MSDA is that, for each query in the current frame, it attends to a subset of key points sampled from multi-scale feature maps of neighboring frames. This allows our method to gather enriched temporal representations and preserve temporal consistency of the foreground instances. The formulation of T-MSDA can be expressed as follows:
\begin{normalsize}
\begin{equation}
\begin{aligned}
\mathcal{T\text{-}MSDA} & (z _{q}, \hat{p} _{q}, x_{t'}^{l}) = \sum_{m=1}^{M}W_{m} \cdot [\sum_{t'=t-d \atop t'!=t}^{t+d}\sum_{l=1}^{L}\sum_{k=1}^{K_{inter}} \\ & A_{mltqk}  \cdot W_{m}^{'} x_{t'}^{l} (\phi_{l}(\hat{p} _{q})+\Delta p_{mtlqk} ) ],
\end{aligned}
\end{equation}
\end{normalsize}
where $t'$ denotes the neighboring frame with frame $t$. $K_{inter}$ represents the number of sampled points in each feature level across neighboring frames.

Let $E_{t}^{intra}$ and $E_{t}^{inter}$ represent the outputs for frame $t$ obtained by $\mathcal{S\text{-}MSDA}$ and $\mathcal{T\text{-}MSDA}$, respectively. According to our findings, $E_{t}^{intra}$ contains more salient instance information about the current frame, which is crucial for distinguishing the instance from the background. On the other hand, $E_{t}^{inter}$ contains more discriminative representations to detect changes in the instance's appearance. To effectively combine the strengths of both modules, we propose a dynamic attention fusion module that uses channel-wise attention to highlight the semantic information. We aggregate the two outputs using element-wise addition and then apply a fully connected layer to adaptively select from $E'$. Subsequently, we use two additional fully connected layers to generate gate vectors $\left \{  g_1,g_2\right \}$, which control the flow of information from the attention maps $\left \{  E_{t}^{intra}, E_{t}^{inter}\right \}$. To generate adaptive weights $\left \{  w_1,w_2\right \}$, we utilize the softmax function along the channel dimension. Finally, we obtain the enriched spatial-temporal features of frame $t$ by performing a weighted addition:
\begin{normalsize}
\begin{equation}
\begin{aligned}
E_{t} = E_{t}^{intra} \odot w_1 + E_{t}^{inter} \odot w_2.
\end{aligned}
\end{equation}
\end{normalsize}



\textbf{ReID embeddings.} To obtain ReID embeddings based on DeformableDETR~\cite{zhu2020deformable}, we introduce a temporal contrastive loss that maps different object queries to unique ReID representations. For each video clip, we consider queries belonging to the same object as positive samples and all other queries as negative samples.

Here define $B_t={\left \{  b_t^q \right \} }_{q=1}^{Q}$ be the object queries of frame $t$, where $b_t^q\in R^C$ and $Q$ is the total number of queries. Automatically, ${\left \{  b_{t}^{i}, b_{t'}^{i} \right \} }$ from frame $t$ and its neighboring frame $t'$ can be treated as the different views of the $i\text{-}th$ instance in terms of temporal domain. We first adopt two FC layers as the ReID head to generate the ReID representations of the queries. Then, InfoNCE~\cite{oord2018representation}~ is employed between two frames, which can be described as:
\begin{normalsize}
\begin{equation}
\begin{aligned}
\mathcal{L}_{N}(B_{t}, B_{t'})=-\frac{1}{Q}\sum_{i}^{Q}log\frac{exp(s(b_{t}^{i}, b_{t'}^{i})/\tau )}{\sum_{j=1}^{Q}exp(s(b_{t}^{i}, b_{t'}^{j})/\tau)},
\end{aligned}
\end{equation}
\end{normalsize} where $\tau$ represents the temperature hyper-parameter, and $s$ denotes the cosine similarity. During inference, the ReID features of the queries are utilized to measure visual similarity. In summary, the total losses of STD-Net can be defined as follows:
\begin{normalsize}
\begin{equation}
\begin{aligned}
\mathcal{L}_{total} = \lambda_{cls}\cdot\mathcal{L}_{cls} + \mathcal{L}_{box} + \mathcal{L}_{cl},
\end{aligned}
\end{equation}
\end{normalsize} where $\mathcal{L}_{cls}$ and $\mathcal{L}_{box}$ represent classification loss and bounding box loss. Herein, we use focal loss~\cite{lin2017focal} as the classification loss. The box loss consists of L1 loss and GIou loss~\cite{rezatofighi2019generalized}: $\mathcal{L}_{box} = \lambda_{L1}\cdot\mathcal{L}_{L1} + \lambda_{giou}\cdot\mathcal{L}_{giou} $.

\begin{table*}[!h]\small
\centering
\renewcommand{\arraystretch}{1.0}
\begin{tabular}{c|ccccc}
\hline
Methods & $\mathcal{J}\&\mathcal{F}$ & $\mathcal{J}$\text{-}$Mean$  & $\mathcal{J}$\text{-}$Recall$ & $\mathcal{F}$\text{-}$Mean$  & $\mathcal{F}$\text{-}$Recall$ \\ \hline
UnOVOST ~\cite{luiten2020unovost}           & 67.9      & 66.4   &76.4   & 69.3  &76.9     \\
Target-Aware~ \cite{zhou2021target}       & 65.0      & 63.7   &71.9    & 66.2   &73.1   \\
Propose-Reduce~ \cite{lin2021video}     & 68.3      & 65.0    &-     & 71.6     &- \\
INO   ~\cite{pan2022n}                  & 72.5      & 68.7    &-     & 76.3     &-  \\
MCMPG + STCN ~\cite{yuan2023}             & 78.4      & 75.4    &83.9  & 81.4 &88.9       \\
\textbf{UVOSAM}                 & \textbf{80.9}      & \textbf{77.5} & \textbf{84.9}    & \textbf{84.2}  & \textbf{91.7} \\ \hline     
\end{tabular}
\vspace{1.3mm}
\caption{The performance comparison of our method and state-of-the-art methods on DAVIS2017-unsupervised validation set. For MCMPG, we showcase its best results combined with the STCN~\cite{Cheng2021}.}
\label{tab1}
\end{table*}

\begin{table*}[!h]\small
\centering
\renewcommand{\arraystretch}{1.0}
\begin{tabular}{cc|ccccc|ccccc}
\hline
\multicolumn{2}{c|}{\multirow{2}{*}{Methods}}                   & \multicolumn{5}{c|}{YoutubeVIS-2019} & \multicolumn{5}{c}{YoutubeVIS-2021} \\ \cline{3-12} 
\multicolumn{2}{c|}{}                                                                &  $\rm AP$    & $\rm AP_{50}$   & $\rm AP_{75}$ & $\rm AR_{1}$   & $\rm AR_{10}$ & $\rm AP$     & $\rm AP_{50}$   & $\rm AP_{75}$  & $\rm AR_{1}$   & $\rm AR_{10}$ \\ \hline
 CrossVIS~\cite{yang2021crossover} &{}                     & 36.3  & 56.8  & 38.9  & 35.6  & 40.7 & 34.2  & 54.4  & 37.9  & 30.4  & 38.2 \\
 VISOLO~\cite{han2022visolo}  &{}                      & 38.6  & 56.3  & 43.7  & 35.7  & 42.5 & 36.9  & 54.7  & 40.2  & 30.6  & 40.9 \\
IFC~\cite{hwang2021video}    &{}                    & 41.2  & 65.1  & 44.6  & 42.3  & 49.6 & 35.2  & 55.9  & 37.7  & 32.6  & 42.9 \\
 Mask2Former-VIS~\cite{cheng2021mask2former}  &{}               & 46.4  & 68.0  & 50.0  & -     & -    & 40.6  & 60.9  & 41.8  & -     & -    \\
 TeViT~\cite{yang2022temporally}     &{}                     & 46.6  & 71.3  & 51.6  & 44.9  & 54.3 & 37.9  & 61.2  & 42.1  & 35.1  & 44.6 \\
 SeqFormer~\cite{wu2021seqformer}     &{}                  & 47.4  & 69.8  & 51.8  & 45.5  & 54.8 & 40.5  & 62.4  & 43.7  & 36.1  & 48.1 \\
 MinVIS~\cite{huangminvis}   &{}                         & 47.4  & 69.0  & 52.1  & 45.7  & 55.7 & 44.2  & 66.0  & 48.1  & 39.2  & 51.7 \\
IDOL~\cite{wu2022defense}  &{}                       & 49.5  & 74.0  & 52.9  & 47.7  & 58.7 & 43.9  & 68.0  & 49.6  & 38.0  & 50.9 \\
 VITA~\cite{heovita}       &{}                      & 49.8  & 72.6  & 54.5  & 49.4  & 61.0 & 45.7  & 67.4  & 49.5  & 40.9  & 53.6 \\
GenVIS~\cite{heo2022generalized}  &{}                       & 51.3  & 72.0  & 54.6  & 49.5  & 59.7 & 47.1  & 67.5  & 51.5  & 41.6  & 54.7 \\
 \textbf{UVOSAM}  &{}                       &  \textbf{52.4}  &  \textbf{74.1}  &  \textbf{56.8}  &  \textbf{51.2}  &  \textbf{61.9} &  \textbf{48.4}  &  \textbf{69.4}  &  \textbf{53.8}  &  \textbf{42.1}  &  \textbf{56.3} \\ \hline
\end{tabular}
\vspace{1.3mm}
\caption{The performance comparison of our method and state-of-the-art methods on Youtube-VIS 2019 and Youtube-VIS 2021 validation sets. UVOSAM performs better than all existing mask-supervised methods with the backbone of ResNet-50.}
\label{tab2}
\end{table*}

\section{Experiment}
\subsection{Experimental Setup}
\textbf{Datasets.} We evaluate UVOSAM on three popular video segmentation datasets: DAVIS2017-unsupervised \cite{pont20172017} and Youtube-VIS 2019\&2021 \cite{yang2019video}. Video instance segmentation builds upon unsupervised video object segmentation by incorporating object category recognition. Therefore, the corresponding dataset of YouTubeVIS can also be used for experiments. The DAVIS2017-unsupervised dataset consists of 120 high-quality videos in total, which are further split into 60 for training, 30 for validation, and 30 for test-dev. The YoutubeVIS-2019 dataset comprises 2,238 training, 302 validation, and 343 test videos. The semantic category number is 40. The YoutubeVIS-2021 dataset adds more samples based on YoutubeVIS-2019, with 3,859 high-quality videos, while the number of annotations is also doubled.

\textbf{Evaluation Metrics.}
For DAVIS2017-unsupervised, we employ the official evaluation measures including region similarity $\mathcal{J}$, boundary accuracy $\mathcal{F}$ and the overall metric $\mathcal{J}\&\mathcal{F}$. For Youtube-VIS dataset, we adopt Average Precision (AP) and Average Recall (AR).

\textbf{Implements Details.}
For STD-Net, we select ResNet-50 \cite{he2016deep} as the default backbone network. All hyper-parameters related to the network architecture are the same as those in the official DeformableDETR. For DAVIS2017-unsupervised, we modify the class head to a binary classification output, identifying only whether the object is foreground. We use 4 NVIDIA A100 GPUs with a global batch size of 16. We first pre-train STD-Net on MSCOCO \cite{lin2014microsoft} for 120,000 steps and then train on DAVIS2017-unsupervised and Youtube-VIS for 2,0000 steps and 80,000 steps, respectively. The AdamW optimizer \cite{DBLP:conf/iclr/LoshchilovH19} is adopted with a base learning rate of $1 \times 10^{-4}$ and $5 \times 10^{-5}$ for the two-stage training. For SAM, we employ the pre-trained model with the ViT-H \cite{dosovitskiy2020transformers} image encoder as the base segmentation model. Following mainstream methods, we resize the frame to a shorter size of 360 and 480 pixels on Youtube-VIS and DAVIS2017-unsupervised during inference, respectively.

\subsection{Main Results}
We compare UVOSAM with the state-of-the-art methods on DAVIS2017-unsupervised and YouTube-VIS 2019 \& 2021. Among these competitive methods, our UVOSAM sets new state-of-the-art results, using only the training data of boxes.

\textbf{DAVIS2017-unsupervised.} \cref{tab1}~illustrates the results of all compared methods on DAVIS2017-unsupervised validation set. Our UVOSAM achieves $\mathcal{J}\&\mathcal{F}$ scores of 80.9$\%$, which outperforms current mask-supervised methods by a large margin. Specifically, UVOSAM makes 2.5 $\%$ absolute improvements over the best competitor MCMPG+STCN, which utilizes a more complex ResNeXt-101 backbone.


\textbf{YoutubeVIS-2019\&2021.} We present our comparisons of YoutubeVIS-2019\&2021 in ~\cref{tab2}. When using the ResNet-50 backbone, our UVOSAM sets a new state-of-the-art with AP of 52.4$\%$ and 48.4$\%$, respectively. The higher AP and AR values demonstrate that our method can effectively leverage enriched spatial-temporal information to optimize segmentation results. Notably, without using any masks for training, our UVOSAM achieves an absolute gain of 1.1$\%$~\&1.3$\%$ AP over the previous best method, GenVIS. Additionally, it is worth mentioning that UVOSAM outperforms GenVIS by approximately 2.3$\%$ in the overlap threshold of $\rm AP_{75}$, indicating that UVOSAM is capable of optimizing the non-overlapping cases.

\begin{figure*}[ht] 
\centering 
\includegraphics[scale=0.58]{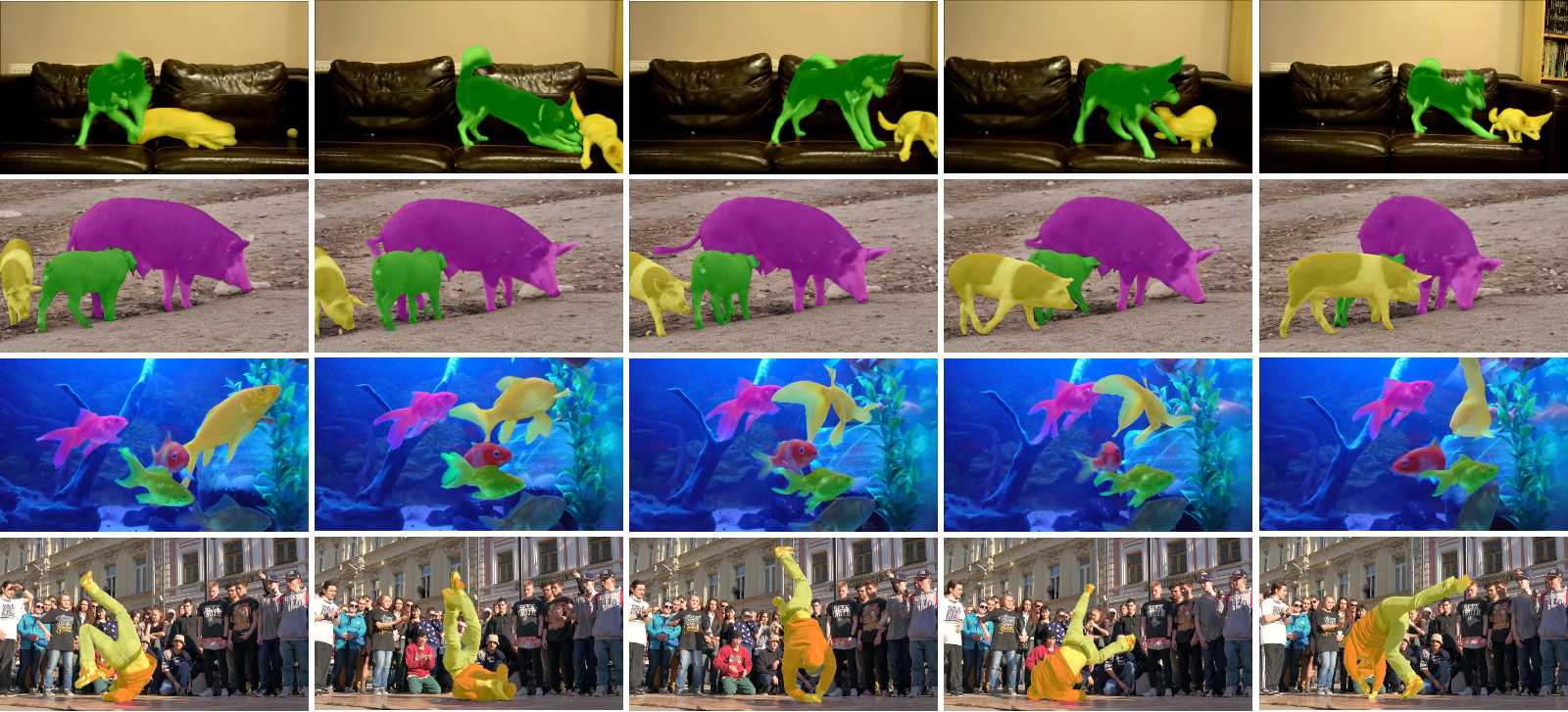} 
\caption{The qualitative results of our proposed UVOSAM on DAVIS2017-unsupervised and Youtube-VIS datasets.} 
\label{fig3} 
\vspace{-3.5mm}
\end{figure*}

 \cref{fig3} visualizes the results of UVOSAM in four challenging scenarios. It shows UVOSAM can generate high quality masks and maintain stable tracking trajectories.

\subsection{Ablation Study}
In this section, we conduct ablation studies to figure out the contribution of different designs. All the ablation experiments are conducted on the DAVIS2017-unsupervised, using the same implementation details as previously described.

\textbf{The impact of the proposed modules on the quality of prompt.} We verify the effectiveness of the proposed modules for improving the quality of box prompts. Firstly, we train the baseline DeepOCSORT~\cite{maggiolino2023deep} with the detector of DeformableDETR. As shown in  ~\cref{tab3}, it only achieves $\mathcal{J}\&\mathcal{F}$ of 78.1 $\%$. Compared to the baseline model, the introduction of temporal contrastive learning provides an obvious gain of 1.0 $\%$. By performing STD-DA in the deformable transformer, it further brings the $\mathcal{J}\&\mathcal{F}$ to 80.9$\%$. In summary, our STD-Net outperforms baseline by 2.8$\%$. Moreover, we additionally offer box annotations as prompts for SAM, which are specifically denoted as "Humans". The superior UVOS results produced by SAM are evident when the prompts are sufficiently accurate.

 \begin{table}[!h]\small
\centering
\renewcommand{\arraystretch}{1.0}
\begin{tabular}{c|ccc}
\hline
Methods & $\mathcal{J}\&\mathcal{F}$ & $\mathcal{J}$\text{-}$Mean$  & $\mathcal{F}$\text{-}$Mean$  \\ \hline
baseline    & 78.1 & 74.5 & 81.7   \\
baseline +TCL & 79.1      & 75.8     & 82.3        \\ 
baseline + TCL + STD-DA & \textbf{80.9}      & \textbf{77.5}     & \textbf{84.2}        \\  \hline
Humans* &  87.3      & 83.5        & 91.0   \\ \hline

\end{tabular}
\vspace{0mm}
\caption{Ablation Study of the effective components of STD-Net for improving the quality of box prompts.}
\label{tab3}
\vspace{-1.0mm}
\end{table}

\textbf{Bidirectional neighboring window size of STD-DA.} In \cref{tab4}, we study the impact of the bidirectional neighboring window size $d$ on overall performance. By increasing $d$ from 1 to 3, it brings the $\mathcal{J}\&\mathcal{F}$ to $79.7\%$ , $80.4\%$ and $80.9\%$, respectively.  It is evident that a larger window size can utilize richer temporal information from multiple frames, which facilitates better understanding of the context of instances. To achieve a balance between computation and precision, we set the value of $d$ to 3 for all experiments. Nonetheless, we hypothesize that our performance will continue to improve as $d$ increases.

\begin{table}[!h]\footnotesize
\centering
\renewcommand{\arraystretch}{1.0}
\scalebox{0.965}{
\begin{tabular}{c|ccccc}
\hline
 $d$ & $\mathcal{J}\&\mathcal{F}$ & $\mathcal{J}$\text{-}$Mean$  & $\mathcal{J}$\text{-}$Recall$ & $\mathcal{F}$\text{-}$Mean$  & $\mathcal{F}$\text{-}$Recall$ \\ \hline
0 &78.8	&75.5	&82.7	&82.0	&89.4 \\
1 &79.7	&76.5	&84.0	&82.9  &90.5  \\
2   &80.4	&77.4	&83.9	&83.4	& 91.1   \\
3   &80.9	&77.5	&84.9	&84.2	& 92.0   \\
4   &81.1	&77.6	&85.2	&84.5	&92.3  \\ \hline   

\end{tabular}
}
\caption{Ablation on the bidirectional window size. The performance increases as the temporal window size increases.}
\label{tab4}
\vspace{-1.0mm}
\end{table}

\textbf{Number of temporal sampling points of STD-DA.} \cref{tab5} shows there is a large performance improvement when $K_{inter}$ changes from 1 to 4. It can be seen that when $K_{inter}$ further increases, the performance is basically unchanged. Hence, we choose $K_{inter} = 4$ as our default settings.

\begin{table}[!h]\footnotesize
\centering
\renewcommand{\arraystretch}{1.0}
\scalebox{0.895}{
\begin{tabular}{c|ccccc}
\hline
 $K_{inter}$ & $\mathcal{J}\&\mathcal{F}$ & $\mathcal{J}$\text{-}$Mean$  & $\mathcal{J}$\text{-}$Recall$ & $\mathcal{F}$\text{-}$Mean$  & $\mathcal{F}$\text{-}$Recall$ \\ \hline
0 &78.8	&75.5	&82.7	&82.0	&89.4 \\
1 &79.3	&75.9	&83.3	&82.6	&89.1   \\
2   &79.9	&76.6	&83.9	&83.1	&90.6   \\
3   &80.4	&77.1	&84.4	&83.7	&91.9    \\
4   &80.9	&77.5	&84.9	&84.2	&92.0  \\
5   &81.0	&77.6	&85.0	&84.4	&92.4 \\
6   &80.8	&77.3	&84.9	&84.2	&92.1  \\ \hline   

\end{tabular}
}
\vspace{0mm}
\caption{Ablation on the temporal sampling points. The performance becomes stable when sampling points reach 4.}
\label{tab5}
\end{table}
\vspace{-1.0mm}

\begin{figure}[t] 
\centering 
\includegraphics[scale=0.45]{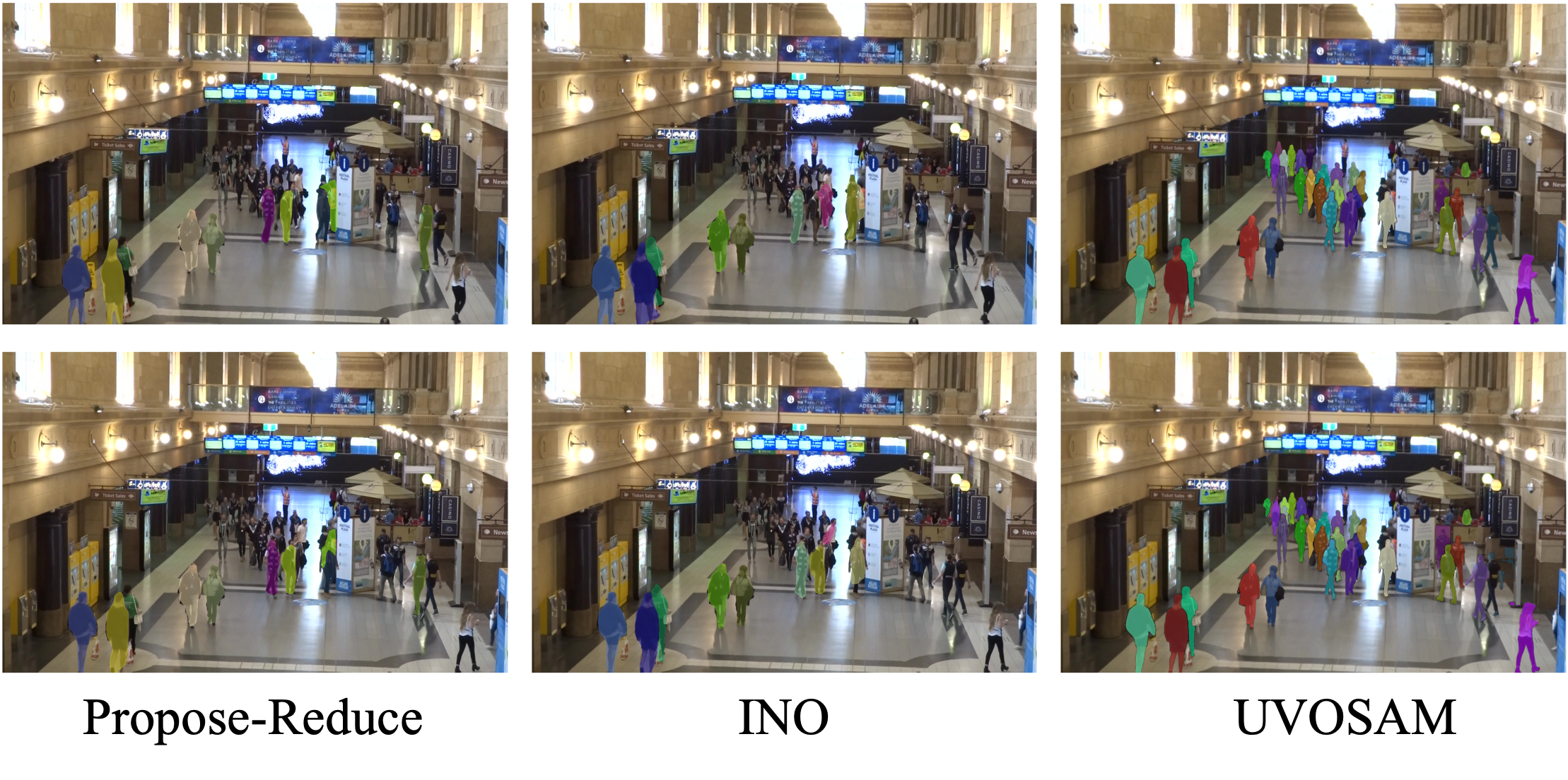} 
\caption{The qualitative comparison of the mask quality on MOT20. 
More visualizations can be found in supplementary materials, best viewed when zoomed in.} 
\label{fig4} 
\vspace{-2.0mm}
\end{figure}

\textbf{Runtime Analysis.} In terms of speed, UVOSAM achieves $11.7$ FPS on DAVIS2017, which is comparable to MCMPG (about 10.5 FPS).

\textbf{The Generalization ability to weakly-annotated video data.} To assess the transfer ability of UVOSAM, we additionally conduct experiments on the MOT20~\cite{dendo20}, which only provides box annotations. As depicted in ~\cref{fig4}, the existing UVOS methods fail to produce satisfactory segmentation results due to the lack of training data. In contrast, UVOSAM, leveraging the weak annotations of boxes, demonstrates exceptional segmentation performance.

\section{Conclusion}
This paper presents UVOSAM, a mask-free framework that utilizes the SAM to address the issue of label-expensiveness in UVOS. To resolve segmentation ambiguity problems caused by off-size box prompts in complex scenes, we introduce a novel tracking network called STD-Net within UVOSAM. This network optimizes non-overlapping bounding boxes using spatial-temporal decoupled deformable attention, and the trajectories generated by it are used as prompts in SAM to sequentially generate segmentation results. We evaluate our approach on three popular datasets and the results show that UVOSAM outperforms mask-supervised methods obviously. Furthermore, our framework demonstrates impressive transferability in handling segmentation on weakly annotated video data. We hope that our work will inspire future research on label-efficient UVOS by promoting the development of fundamental vision models.

\bibliography{aaai24}

\end{document}